\documentclass[manuscript]{acmart}
\AtBeginDocument{%
  }

\usepackage{tabularx}
\usepackage{pifont}
\usepackage{booktabs}
\newcolumntype{Y}{>{\centering\arraybackslash}X}
\newcommand{\cmark}{\textcolor{green!60!black}{\ding{51}}} 
\newcommand{\xmark}{\textcolor{red}{\ding{55}}} 
\newcommand{\pcmark}{(\textcolor{green!60!black}{\ding{51}})}

\setcopyright{none}
\begin{document}

\title{Automated Event Log Generation from Unstructured Text Using Finetuned LLMs}


\author{Maximilian Seeth}
\correspondingauthor
\affiliation{%
  \institution{LMU Munich}
  \city{Munich}
  \country{Germany}
}
\email{Max.Seeth@campus.lmu.de}

\author{Gabriel Marques Tavares}
\affiliation{%
  \institution{LMU Munich}
  \city{Munich}
  \country{Germany}}
\email{tavares@dbs.ifi.lmu.de}

\author{Daniel Schuster}
\affiliation{%
  \institution{University of Mannheim}
  \city{Mannheim}
  \country{Germany}
}
\affiliation{%
  \institution{LMU Munich}
  \city{Munich}
  \country{Germany}
}
\email{daniel.schuster@uni-mannheim.de}

\renewcommand{\shortauthors}{Seeth et al.}

\begin{abstract}
Process mining (PM) provides a powerful framework for discovering and optimizing operational processes from event data. However, the efficacy of PM techniques is strictly predicated on the availability of structured event logs. Thus far, event logs have often been laboriously created by domain and process mining experts. This costly effort causes large portions of organizational knowledge, including incident tickets, manuals, and textual reports, to remain underutilized. We address this bottleneck by investigating the efficacy of Large Language Models (LLMs) as automated data translators. We propose a scalable framework that leverages LLMs as data translators to bridge the gap between unstructured textual resources and structured event data. We finetune LLMs on a newly created text-to-log dataset, demonstrating that the resulting models can extract high-fidelity event logs from unstructured resources. Our results show that this finetuning approach outperforms few-shot or zero-shot prompting  by a large amount, highlighting finetuning as a necessary pre-condition for generating reliable event data. We conclude that our method provides a promising pipeline for making previously unused data available to PM techniques.

\end{abstract}




\maketitle

\section{Introduction} 

Process mining (PM) provides a framework for analyzing and improving operational processes by extracting insights from event data. The success of PM techniques rely heavily on access to well-structured event logs. Thus far, event logs have often been laboriously created by domain and process mining experts. 
Data preparation and extraction are identified as key challenges when applying PM in industrial contexts~\cite{delphi}, as they require substantial time and effort.
This costly effort causes large portions of organizational knowledge, such as incident tickets, manuals, and textual reports, to remain underutilized. We address this bottleneck by investigating the efficacy of Large Language Models (LLMs) as automated data translators. Specifically, we evaluate whether LLMs can reliably map natural language process descriptions into coherent, schema-compliant event logs that can be used by PM.

\begin{figure} 
    \centering \includegraphics[width=\linewidth,clip]
        {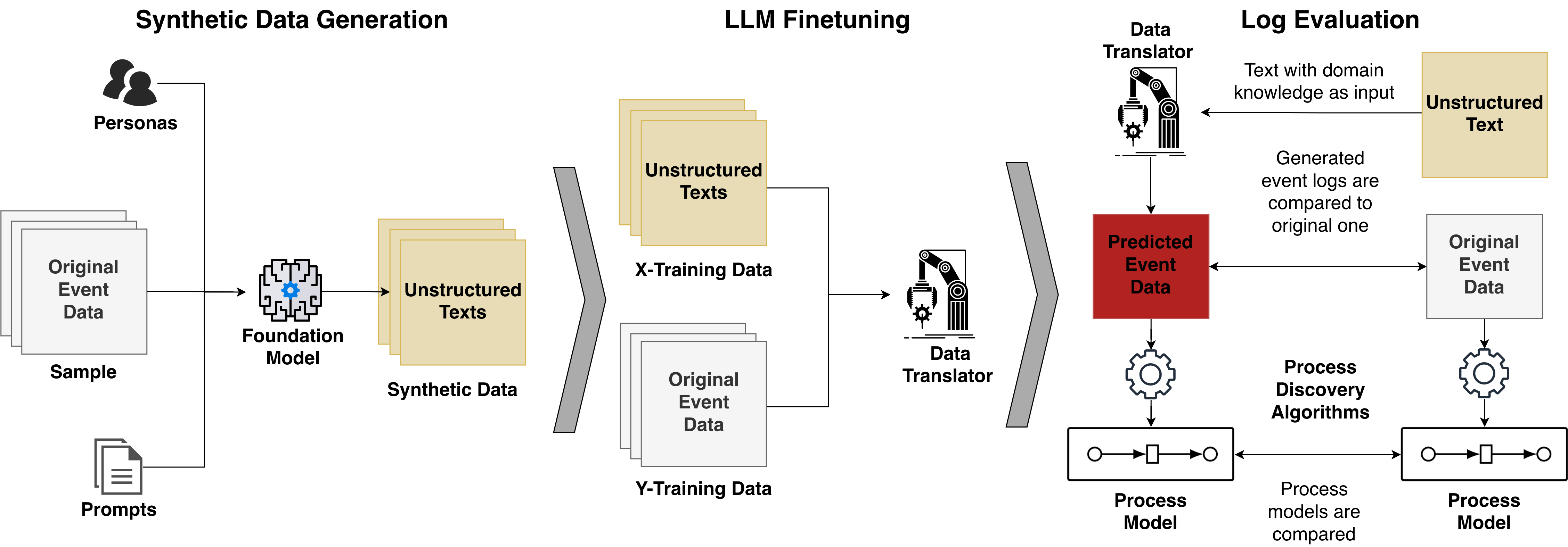} 
  \caption{Pipeline depicting the generation of synthetic texts with frontier models ("Foundation Model" in left panel) that we used to address the lack of a suitable dataset, the finetuning of LLMs ("Data Translator" in centered panel) on this data, as well as the final evaluation of generated event data from texts (right panel) compared to the original data.}
  \label{fig:pipeline}
\end{figure}

Previous attempts to extract process information from unstructured data largely focus on textual resources \cite{Knig2025UnstructuredDI}. We follow this promising tradition \cite{Aa2018ChallengesAO} because tickets, reports, or manuals can be easily collected for further use. Recent advances in LLMs open-up a new paradigm where models go beyond narrow pattern detection and also allow the generation of complex outputs. We investigate these advances to create structured data typically used in PM. This approach differs from other work in PM \cite{Berti2025SpecializingLL,Kourani2024ProcessMW,Lindner2025} that uses LLMs to generate complete process models or has not experimented with supervised finetuning yet \cite{Buss2025,Kecht2021EventLC,SanchezFerriz2025EventLE}. Our idea is to delegate the process modeling to the specified algorithms with the aim of using LLMs to provide them with the appropriate data. Therefore, the guiding research question of our work is \emph{how well LLMs can translate process descriptions to executable event logs}. To make this work more appealing to PM practitioners, we further narrow our experiments to relatively small, and open-weight LLMs that can be deployed locally. 

This paper makes the following key contributions.
(1) An open-weight LLM-based approach for transforming unstructured textual data into structured event logs.
(2) Our results show that only supervised finetuning (SFT), compared to zero-shot and few-shot prompting, substantially improves the ability of LLMs to generate accurate, well-formed event data. This opens a scalable path to producing high-quality event logs from natural language process descriptions.
(3) To arrive at these results, we create synthetic text-to-log datasets since, to the best of our knowledge, there is no suitable dataset that could be used for our goal. We view the use of synthetic data from LLMs as an additional contribution of our work, offering a promising direction for further exploration in PM. 
(4) We introduce an extensive evaluation framework that assesses both individual traces and process-level perspectives, providing a blueprint for future work on generating event logs from unstructured data using LLMs.

The remainder of this paper is structured as follows: Section \ref{Related Work} reviews related research,
Section \ref{Data} presents and evaluates the synthetic data generation, Section \ref{Method} describes our prompting strategies and SFT of LLMs, Section \ref{Evaluation} reports the experimental setup and results\footnote{For reproducibility purposes, we make our training, evaluation scripts, and dataset samples available: \href{https://github.com/omseeth/Automated-Event-Log-Generation-from-Unstructured-Text-Using-Finetuned-LLMs}{\url{https://github.com/omseeth/Automated-Event-Log-Generation-from-Unstructured-Text-Using-Finetuned-LLMs}}}, and Sections \ref{Discussion} and \ref{Conclusion} provide discussion and conclusions.
\begin{table*}[t]
\centering
\footnotesize
\renewcommand{\arraystretch}{1.3} 
\caption{Comparison of different methods using language models (LM) for process mining. \xmark \space indicate that the respective aspect is not used or evaluated and vice versa \cmark \space means it is part of the approach.\space (\cmark) \space mean that the aspect is partially present.}
\begin{tabularx}{\textwidth}{l Y Y Y Y Y Y Y Y Y Y}
\hline
& \textbf{LM type} & \textbf{Dataset} & \textbf{Public data} & \textbf{Prompt variat.} & \textbf{SFT} & \textbf{RL} & \textbf{Log extrac.} & \textbf{Model extrac.} & \textbf{Trace eval.} & \textbf{Process eval.} \\
\hline
\citet{SanchezFerriz2025EventLE} & encoder & unlabeled & \xmark & \xmark & \xmark & \xmark & \cmark & \xmark & \xmark & \xmark \\ 
\citet{Kecht2021EventLC} & encoder & unlabeled & \cmark & \xmark & \xmark & \xmark & \cmark & \cmark & \pcmark & \cmark \\ 
\citet{Buss2025} & decoder & semi-synthetic & \cmark & \xmark & \xmark & \xmark & \cmark & \xmark & \cmark & \xmark \\ 
\citet{Hamdi2025} & decoder & manually labeled & \xmark & \xmark & \xmark & \xmark & \cmark & \xmark & \cmark & \xmark \\ 
\citet{Brennig2025} & decoder & self-supervised & \cmark & \xmark & \cmark & \xmark & \pcmark & \xmark & \cmark & \xmark \\
\citet{Berti2025SpecializingLL} & decoder & manually labeled & \xmark & \xmark & \pcmark & \cmark & \xmark & \cmark & \xmark & \cmark \\ 
\emph{This paper} & decoder & semi-synthetic & \cmark & \cmark & \cmark & \xmark & \cmark & \cmark & \cmark & \cmark \\ \hline
\end{tabularx}
\label{tab:method-comparison}
\end{table*}

\section{Related Work} 
\label{Related Work}

With the advent of a new generation of Transformer based encoder and decoder language models, a growing body of work has explored their application to process mining tasks \cite{Berti2023AbstractionsSA,Berti2025SpecializingLL,Buss2025,Hamdi2025,Grohs2023LargeLM,Kecht2021EventLC,Kourani2024ProcessMW,Lindner2025,Rebmann2024EvaluatingTA,Redis2024ProcessTBenchAL,SanchezFerriz2025EventLE}. One prominent line of research with LLMs investigates prompt engineering techniques to derive process models from textual descriptions \cite{Kourani2024ProcessMW,Lindner2025}. In particular, \citet{Kourani2024ProcessMW} employ prompts and code-level specifications to guide LLMs in generating \texttt{pm4py}~\cite{berti2023pm4py} programs that construct models informed by natural language descriptions of processes. Similarly, \citet{Lindner2025} use prompting as well as retrieval-augmented generation to encourage LLMs to produce structured JSON representations that can be transformed into Dynamic Condition Response (DCR) graphs. However, none of the contributions \cite{Berti2023AbstractionsSA,Grohs2023LargeLM,Kourani2024ProcessMW,Lindner2025,Rebmann2024EvaluatingTA,Redis2024ProcessTBenchAL} have used SFT of LLMs for downstream PM tasks, which is the focus of our approach. Furthermore, most of the prompting based approaches to PM aim for creating full process models directly from unstructured documents.

In contrast, a few notable works focus on extracting logs for PM \cite{Buss2025,Kecht2021EventLC,SanchezFerriz2025EventLE}. \citet{Kecht2021EventLC} and \citet{SanchezFerriz2025EventLE} employ encoder-based language models to derive event representations from unstructured text, which are subsequently synthesized into a coherent event log. For example, \citet{Kecht2021EventLC} leverage the next sentence prediction capabilities of BART: the model receives as input pairs of customer-related inquiries from Twitter together with different ``hypotheses'' indicating to which predefined process activity an inquiry could belong. In this manner, \citet{Kecht2021EventLC} select the most likely activity suggested by the encoder to construct a complete standardized eXtensible Event Stream (XES)~\cite{XES} event log from the textual data. Furthermore, they derive process models from the resulting logs using established discovery algorithms. However, their work differs from ours in three important respects. First, they rely on encoder representations, whereas we focus on decoder-based LLMs. Second, they do not finetune their models for the downstream PM prediction task. Third, their data is unlabeled in the sense that they cannot compare the resulting process models or event logs against any gold standard.

Close to our approach is the work by \citet{Buss2025}. Similarly to our pipeline, \citet{Buss2025} begin by generating synthetic texts based on real-world event logs. Subsequently, they leverage a combination of LLM-based and rule-based NLP techniques to construct event logs from these texts. This process involves first extracting event-related information and then refining it to build an Object-Centric Event Log (OCEL). They deploy GPT-4o mini both for generating the synthetic texts and for extracting process information. Finally, \citet{Buss2025} assess the quality of the resulting traces by comparing them to those from the original logs. The work proposed by \citet{Hamdi2025} is fairly similar to that of \citet{Buss2025}; instead of a synthetic dataset they use a manually labeled one. In contrast to our work, \citet{Buss2025} rely on the same LLM for both text generation and process information extraction. By employing strong foundation models with more nuanced prompting strategies to enhance realism during synthetic text generation, and by using a different set of LLMs to derive event data, we obtain synthetic texts that are substantially more realistic and break a cyclic dependence between the LLMs employed in our approach. Moreover, \citet{Buss2025} and \citet{Hamdi2025} focus their quality analysis exclusively on individual traces. In addition, we derive holistic process representations using discovery algorithms, such as the Inductive Miner and the Heuristics Miner, to evaluate the quality of our logs from a broader process perspective. Finally, neither \citet{Buss2025} nor \citet{Hamdi2025} finetune the LLM in their approach. They neither experiment with different in-context learning strategies (i.e., few- vs zero-shot prompting) as we do.

\citet{Berti2025SpecializingLL} and \citet{Brennig2025} experiment with finetuned LLMs for PM. \citet{Brennig2025} finetune Llama3-8B to predict subsequent events from incomplete traces in XES format. While their approach similarly applies LLMs to structured PM data types, their objective is predicting an event stream continuation; we, however, employ LLMs as complete bridges from a variety of differently formatted natural language texts to structured data. Furthermore, we evaluate the performance of finetuned LLMs also against off-the-shelf ones using different prompting setups. \citet{Berti2025SpecializingLL} specialize differently sized Qwen2.5 models for PM via reinforcement learning (RL) with verifiable rewards, training them to predict footprints of Partially Ordered Workflow Language (POWL) models. Whereas \citet{Berti2025SpecializingLL} treat SFT as an optional warm-up for RL, it is the primary focus of our work where we also consider three different open-weight LLMs (Llama3.1-8B, Ministral-8B, Qwen2.5-7B) with different prompting strategies (zero and few-shot). While \citet{Berti2025SpecializingLL} develop a private dataset of natural language descriptions paired with process models, we focus on generating event logs, not model representations, from natural language artifacts that we publicly share.

A summary of the approaches discussed in this section can be found in Table~\ref{tab:method-comparison}.

\section{Synthetic Data}
\label{Data}

Since existing datasets lack the necessary pairings of natural language process descriptions and structured event logs, we developed two specialized synthetic datasets. We utilize frontier foundation models to transform XES-formatted logs into unstructured textual data.

\subsection{Synthetic Data Generation}
\label{Synthetic Data Generation}

To obtain our main datasets of natural language texts with corresponding event data, we first sample a subset of the \textsc{Road Traffic Fine Management Process} event log \cite{Mannhardt2016Balanced}. The data covers events from issuing invoices to collecting payments and escalating unpaid fines. It comprises approximately 150,000 process traces recorded over a long observation period from January 2000 to June 2012. The event log is characterized by predominantly short traces, with an average of four events per trace, and a large proportion of cases (43\%) terminating after only two events, indicating that many fines were resolved quickly \cite{Mannhardt2016Balanced}. Nevertheless, the log contains 231 variants. We collected a subsample of 2700 traces to gather enough instances for SFT and testing without showing the LLMs too many repetitive examples. To  preserve the original proportions for frequent trace variants we ended up with 2,887 traces because, for infrequent variants, we ensured that at least one instance of each was included.

For our second dataset of natural language texts derived from event data, we sample a subset of the \textsc{Sepsis Cases} event log \cite{MannhardSepsis2017}. The data are derived from sepsis treatments in a Dutch hospital, covering registration, treatment, and discharge of patients. This dataset is much smaller, as it contains overall 1050 traces, and more complex with an average length of 14.49 events per trace as well as 846 different variants. Our sample contains only a subset of 400 traces, effectively cutting-off many variants to keep the test set sizes of samples between the logs comparable.

We split our samples into three parts each, serving as the basis for generating synthetic police reports and medical discharge summaries with three frontier foundation models (Gpt-5.1-2025-11-13, Qwen3-max-2025-09-23, Mistral-medium-2508, all accessed via their respective APIs). We used these models to maximize the quality of the generated output. For each synthetic text generation, the foundation models were presented with varying roles, prompts, and traces (cf.\ Fig.~\ref{fig:pipeline}). To increase textual diversity and realism \cite{Chen2024FromPT}, we define five distinct personas differing in gender, ethnicity, occupation, personality traits, and location, along with three prompts featuring different task descriptions for each log 
(cf.\ repository for details).
Additionally, we employ a stochastic decoding strategy with randomly sampled temperatures ranging from $0.6$ to $0.9$ and a fixed top-p parameter of $0.9$ to introduce a degree of noise into the data.

\subsection{Quality of Synthetic Reports}
\label{Quality of Synthetic Reports}
To assess the quality of our synthetically generated texts, we consider three complementary statistics: (i) document length (ii) \emph{$n$}-gram diversity ratios, and (iii) average pairwise embedding distances. The latter two measures are commonly used as proxies for textual diversity and have been shown to correlate positively with it, as reported in~\cite{Li2025FromMT}. Finally, (iv) we annotate the resulting texts to measure semantic fidelity and temporal clarity (i.e., correct representation of dates and events) w.r.t the log used for their generation.

The \emph{$n$}-gram ratio (ii) quantifies lexical diversity by comparing the number of distinct \emph{$n$}-grams to the total number of observed \emph{$n$}-grams across a corpus. Formally, for a fixed $n$, it is defined as
$$
    R_n
    =
    \frac{\#\text{unique } n\text{-grams}}{\#\text{total } n\text{-grams}}\,.
$$
To capture semantic diversity, we compute average pairwise distances (iii) between document embeddings. Let $\mathbf{v}_a \in \mathbb{R}^d$ denote the embedding of document $a$, and let $N$ be the total number of documents. The average embedding distance is then given by
$$
    D_{\text{avg}} = \frac{1}{N} \sum_{a \neq b} \lVert \mathbf{v}_a - \mathbf{v}_b \rVert\ ,
$$
where $\lVert \cdot \rVert$ denotes the Euclidean norm. Larger values of $D_{\text{avg}}$ indicate greater semantic variability among the generated documents.

\begin{figure*}[t]
\centering
\begin{minipage}[t]{0.49\textwidth}
  \vspace{0pt} 
  \centering
  \includegraphics[width=\linewidth]{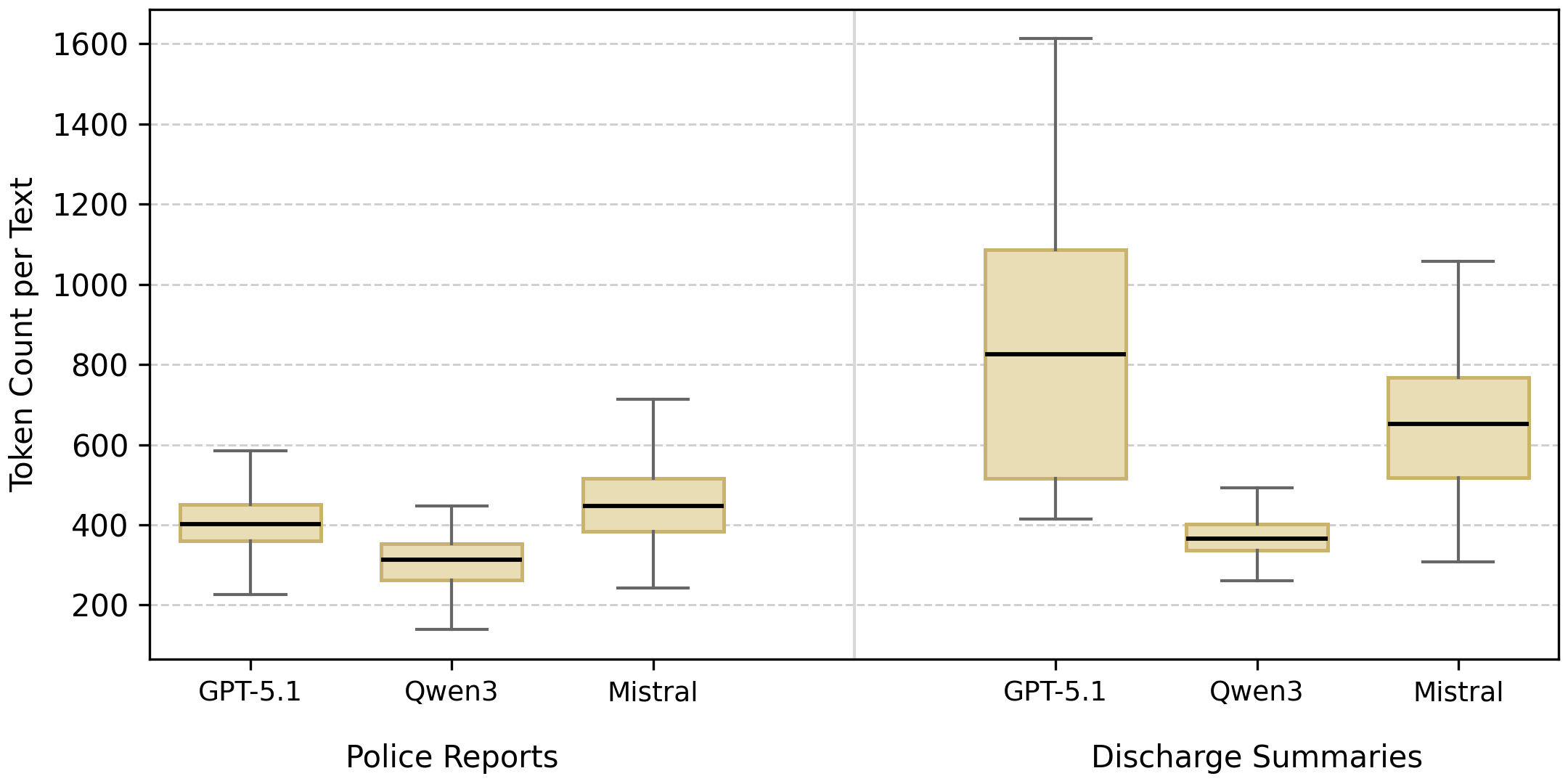}
  \captionof{figure}{Token counts of generated texts with box plots. Left: synthetic
  Police Reports of road traffic fining; right: synthetic Discharge Summaries after
  sepsis treatment in hospitals.}
  \label{fig:synth_token_counts}

  \vspace{\floatsep}
  \footnotesize
  \setlength{\tabcolsep}{4pt}
  \captionof{table}{Average pairwise Euclidean distance between document embeddings for
  each text collection. Higher values imply greater semantic differences.}
  \label{tab:pairwise_distances}
  \begin{tabular}{lc}
  \toprule
  \textbf{Text collection} & \textbf{Avg. pairwise distance} \\
  \midrule
  GPT-5.1 Police Reports & 0.526 \\
  Qwen3 Police Reports & 0.541 \\
  Mistral Police Reports & 0.565 \\
  GPT-5.1 Discharge Summaries & 0.495 \\
  Qwen3 Discharge Summaries & 0.478 \\
  Mistral Discharge Summaries & 0.505 \\
  \bottomrule
  \end{tabular}
\end{minipage}\hfill
\begin{minipage}[t]{0.49\textwidth}
  \vspace{0pt}
  \centering
  \includegraphics[width=\linewidth]{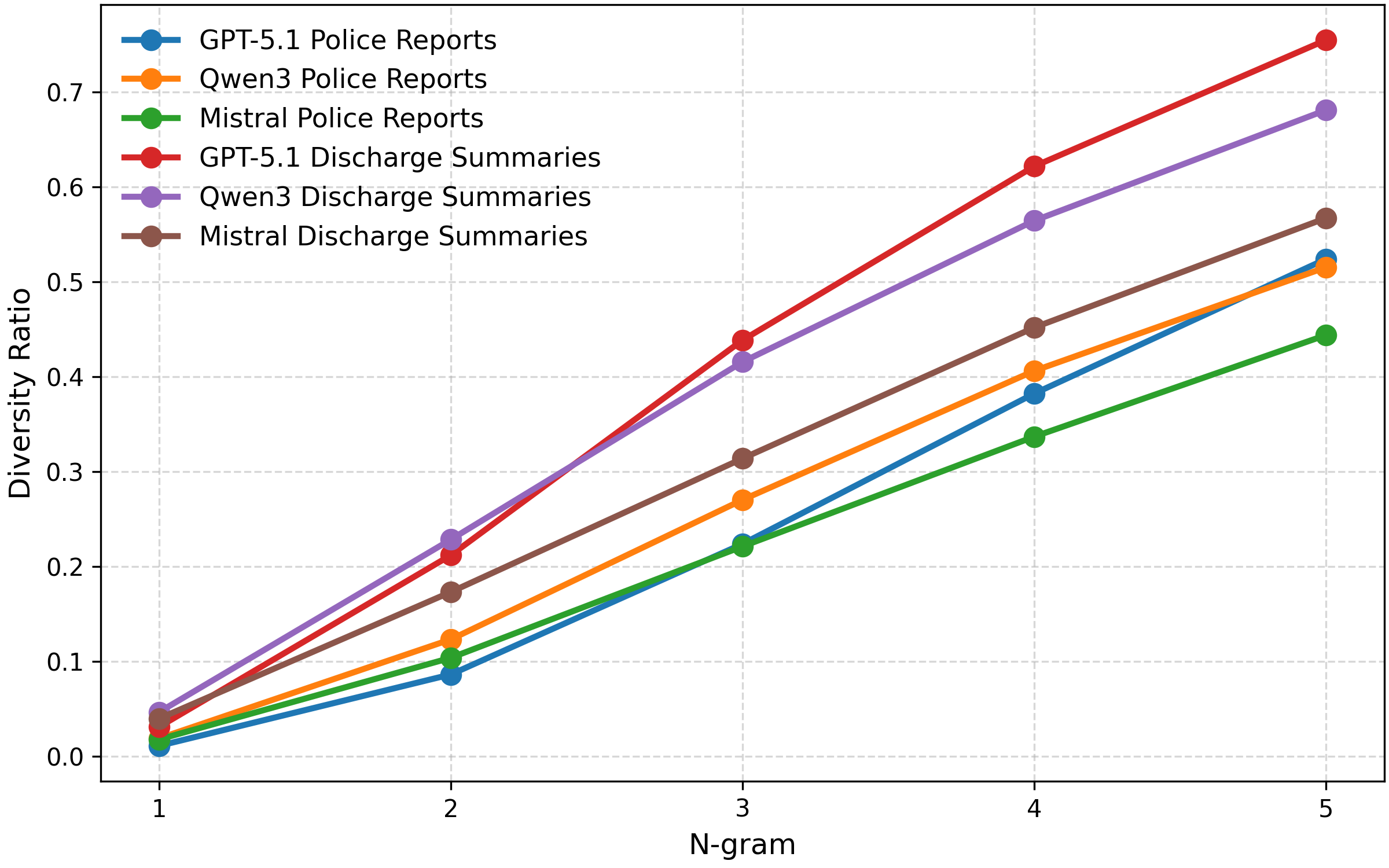}
  \captionof{figure}{$n$-gram ratios w.r.t. all documents. Higher scores imply greater
  textual diversity.}
  \label{fig:synth_diversity}

  \vspace{\floatsep}
  \footnotesize
  \setlength{\tabcolsep}{4pt}
  \captionof{table}{Faithfulness and temporal clarity of synthetic police reports: mean,
  Cohen's Kappa $\kappa$ and percent agreement from two annotators that evaluated 60
  texts on a binary scale.}
  \label{tab:annotation}
  \begin{tabular}{lcc}
  \toprule
  \textbf{Metric} & \textbf{Faithfulness} & \textbf{Temporal clarity} \\
  \midrule
  Mean $\mu$ & 0.950 & 0.983 \\
  Cohen's Kappa $\kappa$ & $-0.029$ & 0.000 \\
  Percent Agreement & 0.900 & 0.967 \\
  \bottomrule
  \end{tabular}
\end{minipage}
\end{figure*}

We report the distribution of document lengths in Fig.~\ref{fig:synth_token_counts}. Across all foundation models, the median remains below 500 tokens for the police reports. This outcome is partly due to our prompts, which encouraged concise answers, and partly reflects the generally short length of the underlying traces used for generation, with only a few longer exceptions. For all reports, the $n$-gram ratio increases monotonically with $n$, indicating a consistent lexical diversity. 

The creation of synthetic discharge summaries led to a much greater variety of texts: the median text length of generated summaries with Gpt-5.1-2025-11-13 is above 800 tokens, whereas the text generated by Qwen3-max-2025-09-23 have less than half the token size. Lengths of texts by Mistral-medium-2508 range in the middle of the other two foundation models. Also the diversity of vocabulary used in the summaries varies across foundation models. Considering 5-grams, Gpt-5.1-2025-11-13 generates more lexical diversity ($R_n=0.75$) than Mistral-medium-2508 ($R_n=0.57$). We attribute the differences to at least two factors: 1) the models received many different trace variants from which they had to generate texts, 2) certain tendencies between the foundation model's tendency to produce different text lengths can be observed where, e.g., Qwen3-max-2025-09-23 always generated the shortest texts, irrespective of the datasets.

To quantify semantic diversity, we compute the average pairwise Euclidean distance between document embeddings within each group of texts generated by the foundation models. We encode each report with the bidirectional encoder model Mxbai-embed-large which has 334M parameters and a context window of 512 tokens from \cite{Lee-embed2024mxbai}. The results, reported in Table~\ref{tab:pairwise_distances}, show that texts generated by Mistral-medium-2508 always have the greatest semantic differences, although results are overall close to each other. Surprisingly, the much more complex discharge summaries are considered to be more similar w.r.t each other than the synthetic police reports, which cover less diverse events. One explanation could be that the embedding model is encoding information from sepsis treatments in a more coarse grained fashion than from road traffic processes due to its underlying pre-training and finetuning.

In addition to the diversity of the generated texts, we evaluate their semantic faithfulness and temporal clarity w.r.t the underlying event log traces from which they were derived. To this end, we randomly sampled 60 synthetic texts from the police report corpus (20 per foundation model). Two annotators independently assessed each text on a binary scale according to the following criteria: (1) whether the synthetic text is faithful to the event order of its trace, i.e., all activities in the trace are present and the narrative flow follows the same order; and (2) whether each activity is clearly associated with a date and whether the dates are consistent with the order of activities. We provide the full annotation guidelines alongside our code. We report the mean annotation scores for criteria (1) and (2), Cohen’s Kappa $\kappa$ for inter-annotator agreement \cite{Cohen1960ACO} 
as well as the raw percentage agreement.

On average, the annotators assigned high scores for both faithfulness and temporal clarity of the synthetic texts ($\mu >= 0.95$; see Table~\ref{tab:annotation}). Despite the high raw agreement (0.900 for faithfulness and 0.967 for temporal clarity), the corresponding Cohen’s Kappa values are low ($\kappa = -0.029$ and $\kappa = 0.000$). This apparent discrepancy can be attributed to the Cohen’s Kappa paradox \cite{Warrens2010AFP}: the base rates of defects (i.e., 0 labels) were low and highly skewed, while overall agreement was very high. Under such conditions, $\kappa$ tends to penalize high agreement, resulting in these deceptively low values. Overall, we report that the generated reports cover in almost all texts the activities and dates from the original log faithfully.

All foundation models used for synthetic text generation yield rather similar quantitative results (consider n-gram ratio in Fig. \ref{fig:synth_diversity} and pairwise distances in Table \ref{tab:pairwise_distances}) for the police reports. Qualitatively, however, we observe several stylistic differences: for example, Gpt-5.1-2025-11-13, tends to redact employee names, whereas Qwen3-max-2025-09-23 and Mistral-medium-2508 more frequently invent names or addresses.

\section{Method}
\label{Method}

\textbf{Zero-Shot and Few-Shot Prompting:} We use zero-shot and few-shot prompting to generate outputs from the LLMs. In both settings, we employ an identical task instruction to ensure comparability. For few-shot prompting, we include a demonstration consisting of a natural language text and XES trace pair. Demonstrations are randomly sampled to mitigate bias arising from repeated exposure to a fixed example \cite{zheng_bias} and to reduce the risk of overfitting to specific demonstration artifacts.

\textbf{Supervised Finetuning of LLMs:} For our finetuning on the police report dataset, we split the generated reports and corresponding traces into training (85\%), development (5\%), and test (10\%) sets. Reports serve as inputs and XES formatted traces as targets for finetuning three instruction-tuned LLMs (consider Fig.~\ref{fig:pipeline}): Llama3.1-8B-Instruct \cite{dubey2024llama3}, Ministral-8B-Instruct-2410 \cite{mistral_ministraux_2024}, and Qwen2.5-7B-Instruct \cite{Yang2024Qwen25TR}, all obtained from HuggingFace. We chose these relatively small models with PM practitioners in mind who would like to deploy this pipeline locally. To allow training with a single GPU, we apply Low-Rank Adaptation (LoRA) with rank 16, updating only a small subset of parameters \cite{Hu2021LoRALA}. The training loss is computed solely over the assistant’s expected output tokens (target), with prompt tokens masked. Each training example includes system prompt, instruction, one-shot demonstration sampled from the training set as well as the desired input-target pair. We share hyperparameter details with the repository.
\section{Experimental Setup and Evaluation}
\label{Evaluation}

\subsection{Experimental Setup}
\label{Experimental Setup}
For the main evaluation, we provide reports from the test set of our police corpus as input to LLMs and prompt them to generate event traces in the XES format of the processes described in natural language. Each inference setup consists of a system prompt with role-based instructions, an explicit specification of admissible activities and attributes, and the report itself, guiding the LLMs during inference. Supplying the allowed activity and attribute set as domain knowledge further allows us to identify hallucinations when the model outputs attributes or activities outside this set. For few-shot prompting, we include only one report–trace pair as an in-context demonstration. This demonstration is sampled from the development set to avoid any leakage from the test set. Following \cite{Modarressi2025NoLiMaLE}, who report performance degradation for inputs exceeding approximately 1{,}000 tokens, we restrict few-shot prompting to one demonstration. Inference is conducted using stochastic decoding (temperature $0.7$, top-$p$ $0.9$).

To evaluate the generalizability of the LLMs, we repeat the inference steps described above on our synthetic discharge summaries, which we likewise split into a development set for demonstration pairs and a test set. We additionally run a held-out experiment in which the LLMs are trained on only one third of the XES traces used in the main experiment, paired exclusively with texts generated by GPT-5.1, and evaluated on a test set of reports generated by Qwen3.

\subsection{Trace Level Evaluation}
\label{Trace Level Evaluation}

\begin{figure}[t]
    \centering
    \begin{minipage}{0.49\textwidth}
        \centering
        \includegraphics[width=.9\linewidth]{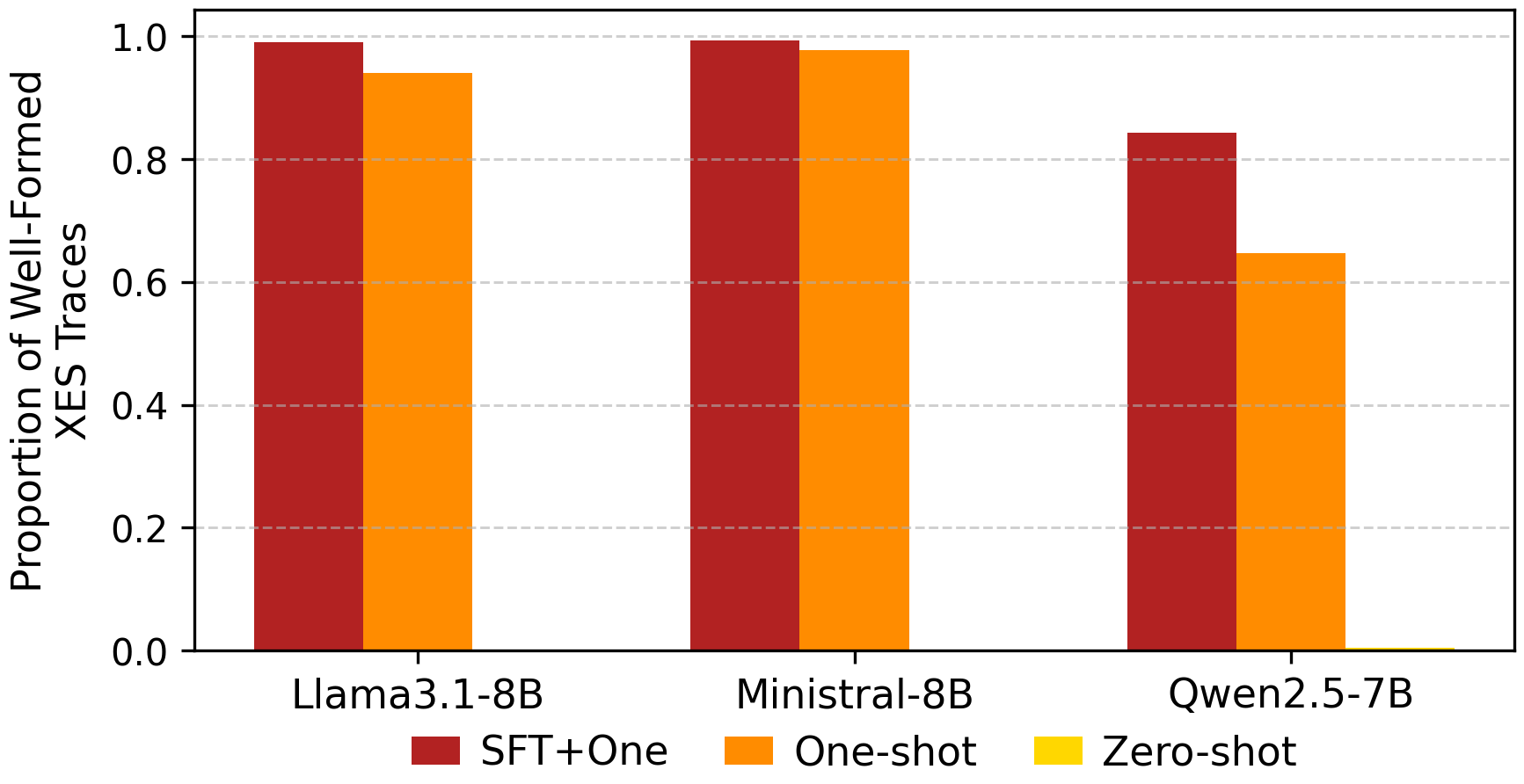}
        \captionof{figure}{Proportion of well-formed traces. SFT and one-shot prompting lead to many valid XES structured outputs from our corpus of synthetic police reports.}
        \label{fig:well_formed_XML}
    \end{minipage}
    \hfill
    \begin{minipage}{0.49\textwidth}
        \centering
        \includegraphics[width=.9\linewidth]{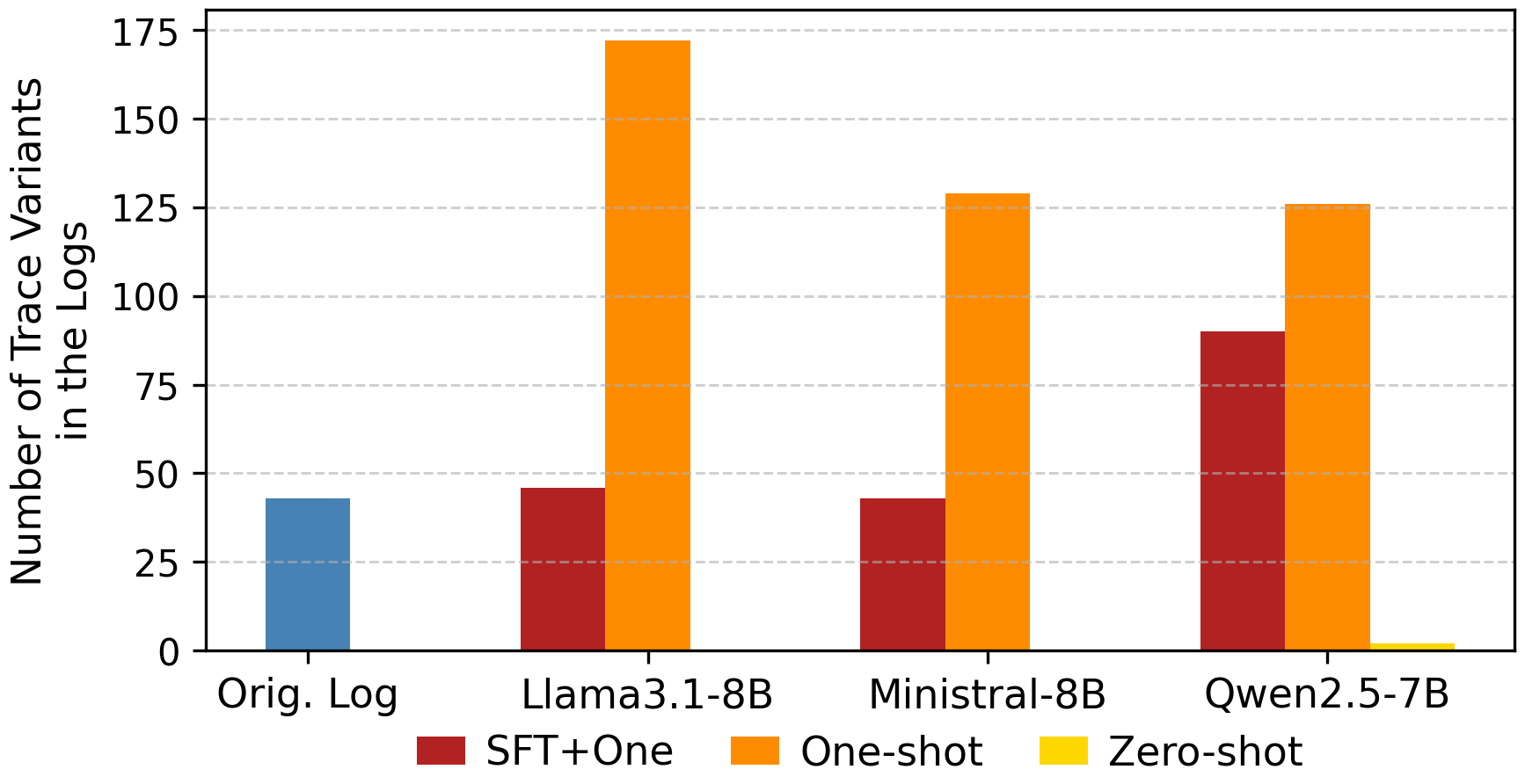}
        \captionof{figure}{Number of variants in the logs for \textsc{Road Traffic Fine Management}. Only SFT roughly aligns the number of variants with the expected number from the original event log.}
        \label{fig:number_variants}
    \end{minipage}
\end{figure}

We consider the number of predicted traces that conform to the XML syntax required for XES data and have complete activities and time-stamps (i.e., which are 'well-formed'), divided by the total number of traces in the test set (see Fig.~\ref{fig:well_formed_XML}). The generated XES structures improve substantially when the LLMs are provided with one-shot demonstrations of police reports with corresponding traces. Under this setting, the models consistently produce well-formed XML-styled XES data points. Finetuning further enhances performance: for instance, the finetuned Llama3.1-8B-Instruct model generated 430 valid traces out of 434. Notably, the Qwen2.5-7B-Instruct model also produced 2 well-formed XES structures in the zero-shot setting, suggesting that it may have been exposed to XES–like syntax during pretraining.

\begin{table}[t]
    \footnotesize
    \setlength{\tabcolsep}{4pt}
    \begin{minipage}[t]{0.49\textwidth}
    \centering
    \caption{Hallucination rate of attributes within valid traces compared across three methods: SFT + One-shot, One-shot and Zero-shot prompting from our synthetic \textsc{Road Traffic Fine Management} logs. "NA" means no valid traces are available.}
    \begin{tabular}{ l l l l }
    \hline
     & \textbf{Llama3.1-8B} & \textbf{Ministral-8B} & \textbf{Qwen2.5-7B} \\
    \hline
    SFT + One-shot & 0.5\% & 0.5\% & 7.6\% \\
    One-shot & 0\% & 0\% & 1.4\% \\
    Zero-shot & NA & NA & 150\% \\
    \hline
    \end{tabular}
    \label{tab:hallucination_attrs}
    \end{minipage}%
    \hfill
    \begin{minipage}[t]{0.49\textwidth}
    \centering
    \caption{Hallucination rate of activities within valid traces compared across three methods: SFT + One-shot, One-shot and Zero-shot prompting from our synthetic \textsc{Road Traffic Fine Management} logs. "NA" means no valid traces are available.}
    \begin{tabular}{ l l l l }
    \hline
     & \textbf{Llama3.1-8B} & \textbf{Ministral-8B} & \textbf{Qwen2.5-7B} \\
    \hline
    SFT + One-shot & 0\% & 0\% & 0\% \\
    One-shot & 0\% & 0\% & 0\% \\
    Zero-shot & NA & NA & 0\% \\
    \hline
    \end{tabular}
    \label{tab:hallucination_activities}
    \end{minipage}
\end{table}

The number of variants in the event logs aligns closely with the expected number of variants of the sample of the original log ($n=43$) after finetuning for Llama3.1-8B-Instruct ($n=46$) and Ministral-8B-Instruct-2410 ($n=43$) with (consider Fig.~\ref{fig:number_variants}). These two models produce far more variants ($n\geq129$) without training. In contrast, Qwen2.5-7B-Instruct generates less variety of different traces ($n=126$) when provided with a demonstration, but still substantially more ($n=90$) than the number of variants in the original log after finetuning.

Providing the LLMs with domain knowledge about activities and attributes leads to low number of hallucinated attributes within well-formed traces for all three models. After finetuning, Llama3.1-8B-Instruct hallucinates only 2 attributes (up from 1 in the one-shot approach). Ministral-8B-Instruct-2410 is also quasi hallucination free. Qwen2.5-7B-Instruct's numbers must be put into perspective as we consider the hallucinations only within those traces that are valid. For Qwen2.5-7B-Instruct 4 hallucinated activities occur in 281 traces for the one-shot approach, and 28 hallucinations occur in 366 traces after finetuning. As for the zero-shot setting, Qwen2.5-7B-Instruct hallucinated three attributes in two traces (i.e., its hallucination rate is 150\%). Consider the rate of hallucinated attributes in all valid traces in Table~\ref{tab:hallucination_attrs}. As for hallucinated activities, the instruction given to the models with the respective domain information leads to no hallucinated activities in well-formed and complete traces (Table~\ref{tab:hallucination_activities}).

Finally, we compare each generated trace with the original one of each respective report. For each comparison of traces, we compute the Levenshtein distance and Kendall's Tau similarity for activities (i.e., event order and event lengths per trace) and overlaps for categorical and numerical attributes. We average and normalize the results for easy comparison.

\begin{figure}[t]
    \centering 
    \includegraphics[width=\linewidth]{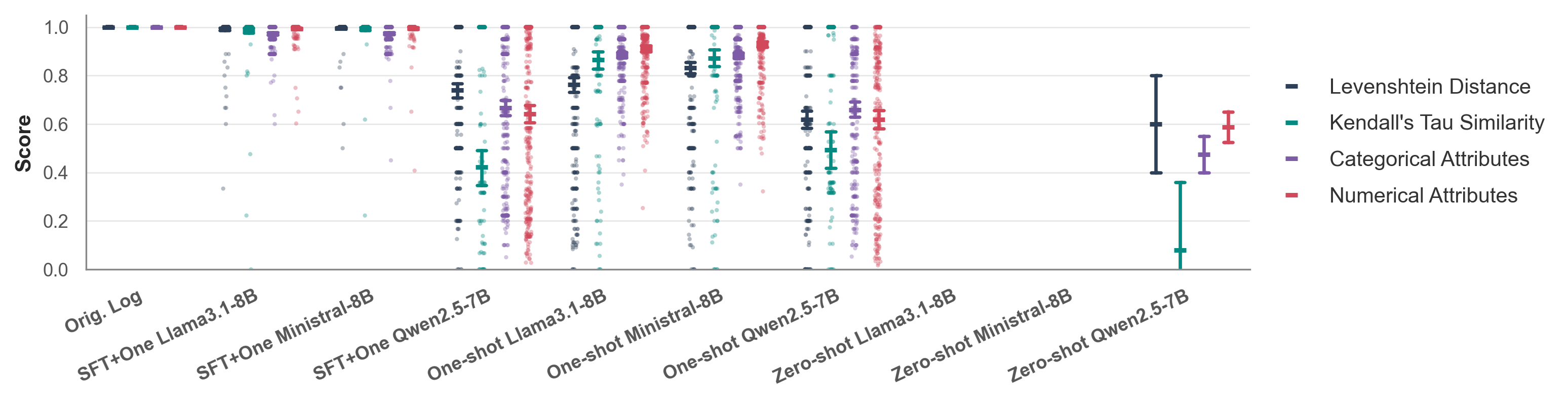} 
        \caption{Normalized metrics (Levenshtein distance, Kendall's Tau similarity, categorical attributes, numerical attributes) for evaluating the quality of individual traces from the generated event logs of our synthetic \textsc{Road Traffic Fine Management} corpora. Each bold point represents the respective metric per trace, with 95\% confidence intervals of the mean. Methods (SFT, one-shot and zero-shot prompting) are indicated as prefixes to the model names. Since no valid event logs were generated with the zero-shot approach, we ignore this method in this figure. SFT improves the generation of traces compared to one-shot prompting setups.}
    \label{fig:trace_level_metrics}
\end{figure}

\textbf{Kendall's Tau:} We measure the similarity between two rankings $\sigma_1$ and $\sigma_2$ using Kendall’s Tau correlation coefficient \cite{KendallTau}. Since the rankings must be of same length, we always trim or pad the sequence of events for a given trace from the predicted log to the length of the trace of the original log. Kendall’s $\tau$ compares the number of concordant and discordant pairs between the rankings and is defined as
$
    \tau(\sigma_1, \sigma_2) = \frac{C - D}{C + D},
$
where $C$ is the number of concordant pairs of event activities and $D$ is the number of discordant pairs. The coefficient takes values in $[-1, 1]$, with $\tau = 1$ indicating identical rankings, $\tau = 0$ no association, and $\tau = -1$ a complete reversal.
Given $m$ pairs of rankings, we compute the Kendall’s tau mean as
$
    \tau_{\mu} = \frac{1}{m} \sum_{k=1}^{m} \tau(\sigma_{k,1}, \sigma_{k,2}).
$

\textbf{Levenshtein distance:} Let $x$ and $y$ be two sequences of events. The Levenshtein distance $d(x,y)$ is the minimum number of insertions, deletions, or substitutions of events (i.e., activities) required to transform $x$ into $y$ \cite{Levenshtein1965BinaryCC}. We normalize this value as
$
    s(x,y) = 1 - \frac{d(x,y)}{\max(|x|,|y|)},
$
where $|\cdot|$ denotes the trace length. The reported score is the mean normalized Levenshtein distance over all traces.

\textbf{Categorical Attributes:} The categorical similarity is defined as:
$
    s_{cat}={\#\text{Correctly predicted categorical attribute values}\over\# \text{Total of categorical attributes in gold trace}}
$
per trace where we only consider the prediction of attribute values for existing attribute values from the original trace. We further average over all traces. This is only a rough estimate.

\textbf{Numerical Attributes:} The numerical similarity is defined as:
$
    s_{num}={\sum \exp(\text{-dist(}v, v'))\over\# \text{Total of numerical attributes}}
$
for given attribute values $v, v'$ per trace where we only consider the prediction of attribute values for existing attribute values from the original trace. We further average over all similarities of the log.

In line with the previous analysis, finetuning substantially improves the quality of the traces generated by the LLMs (see Fig.~\ref{fig:trace_level_metrics}). For Llama3.1-8B-Instruct and Ministral-8B-Instruct-2410, both the ordering and the length of events per trace closely match those of the original event log, as reflected by their high Levenshtein distance (mean of 0.99 for Llama and 1.0 for Ministral) and Kendall's Tau scores (mean of 0.99 for both models). In contrast, the finetuned Qwen2.5-7B-Instruct achieves a stronger Levenshtein distance score (mean of 0.74) than Kendall's Tau similarity (mean of 0.42), suggesting that while the generated traces are largely similar in content, they may omit one or more events that disrupt the global ordering. On average, considering the traces both in the original log and the log from the finetuned Qwen2.5-7B-Instruct model, original traces have a length of 3.95 events, whereas the traces generated with Qwen2.5-7B-Instruct have a length of 3.09 events. Finally, the one-shot setting exhibits substantially larger deviations: while the predicted number of events per trace is comparable to that of the original data, the ordering of events diverges more frequently from the expected log structure, leading to mean values like 0.76 for the Levenshtein distance from Llama3.1-8B-Instruct, and greater discrepancies are also observed in both categorical and numerical attributes compared to the original traces. For example, with one-shot prompting, Qwen2.5-7B-Instruct generates reports whose events overlap with the numerical and categorical attributes of the original log in only about 62\% to 66\% of times per trace.

\subsection{Process Level Evaluation}
\label{Model Level Evaluation}

From a holistic perspective, we evaluate the quality of the generated traces forming complete event logs by discovering process models -- serving as process representations -- using the Inductive Miner \cite{Leemans} and Heuristics Miner \cite{Weijters2006} algorithms, and comparing them with the target process models obtained from the original event logs (see Fig.~\ref{fig:pipeline}). In other words, we investigate if the generated traces derived from the text inputs resemble log features leading to the same process representations as the corresponding non-textual trace data. Process model quality is quantified using fitness, precision, and F1 scores (i.e., the harmonic mean of fitness and precision), computed with respect to the original traces. In PM, fitness measures how well a discovered process model can reproduce the observed event log, i.e., the proportion of recorded behavior that is allowed by the model. Precision measures how much extra behavior the model allows beyond what is observed in the log, penalizing overly general models. 

\begin{figure} 
    \centering 
    \includegraphics[width=0.85\linewidth]
        {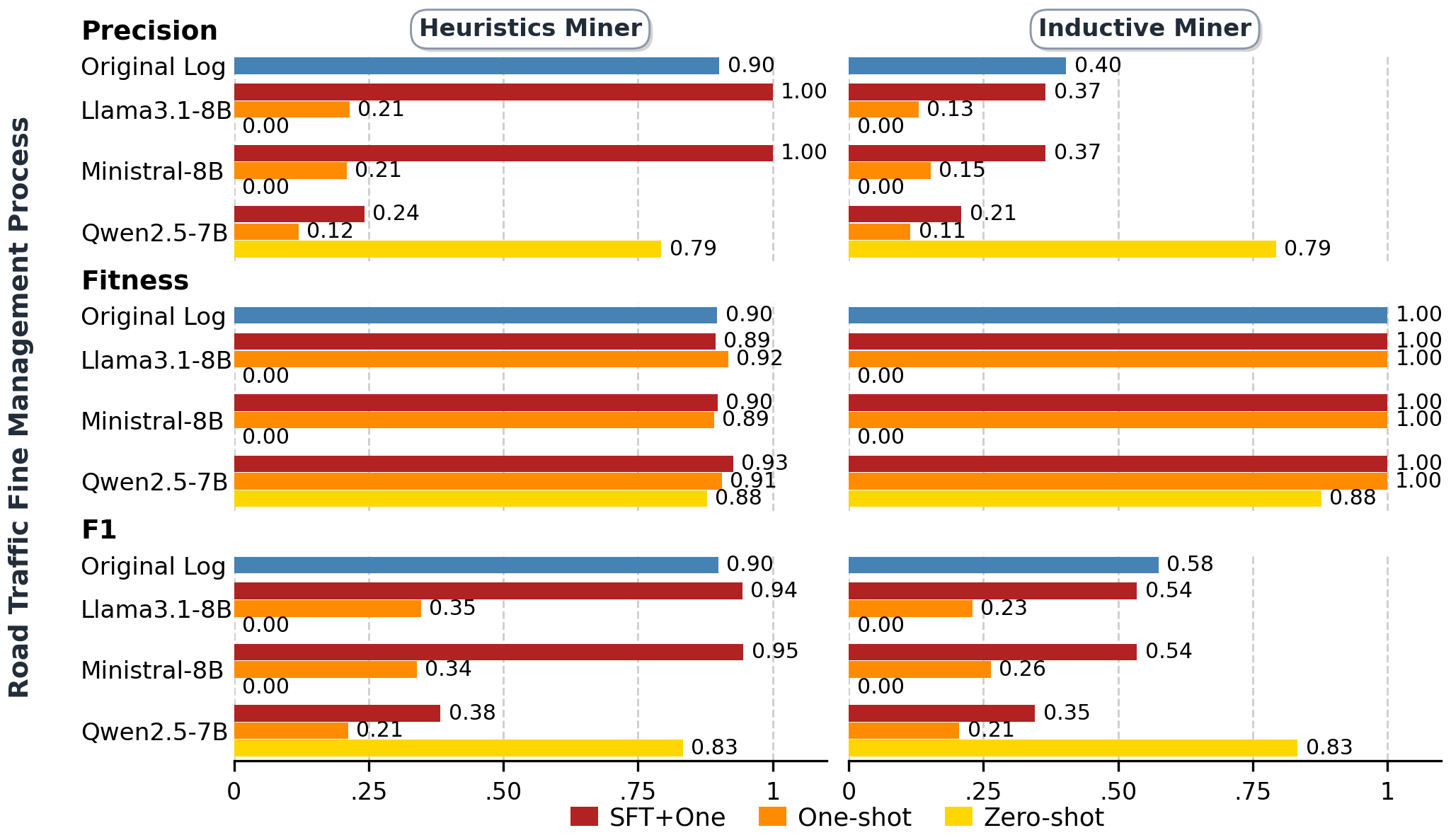} 
        \caption{Precision, Fitness, and F1 scores of Inductive and Heuristics process models w.r.t. the original event log of \textsc{Road Traffic Fine Management}. On average, supervised finetuning (SFT) increases the event log quality from which the process models were derived, compared to one- or zero-shotting the LLMs to generate the event logs from the road traffic fine text data.}
    \label{fig:process_level_metrics_road_traffic_finement}
\end{figure}

For the Inductive Miner, all generated \textsc{Road Traffic Fine Management} logs yield perfect fitness (cf. Fig.~\ref{fig:process_level_metrics_road_traffic_finement}), irrespective of whether the models stem from finetuned or one-shot-generated logs, as the miner is optimized for high fitness. Differences instead appear in precision: while SFT improves precision over one-shot prompting across all logs, none reaches the level of the original log (0.40), since the generated logs contain unfaithful
behavior (cf. the differences in variants in Fig.~\ref{fig:number_variants}). With the Heuristics Miner, fitness likewise remains high throughout, and precision closely matches the original log (0.90) whenever the underlying data is generated by finetuned LLMs. Llama3.1-8B-Instruct and Ministral-8B-Instruct-2410 even attain 1.0, suggesting that the miner removed infrequent variants of the original log and thus produced a better-aligned model. The finetuned Qwen2.5-7B-Instruct is the exception at 0.24, less than one third of the expected precision; its zero-shot counterpart yielded a process model based on only two traces, which explains its outlier scores here and in the following evaluation.

F1 scores of inductively mined models from the generated logs yield a more balanced assessment of event log quality (see Fig.~\ref{fig:process_level_metrics_road_traffic_finement}). Finetuning substantially narrows the gap to the F1 score (0.58) of the real event logs. The process models achieve F1 scores of 0.51 with data from the finetuned Llama3.1-8B-Instruct, 0.44 with data from Ministral-8B-Instruct-2410, and 0.43 with data from Qwen2.5-7B-Instruct. A similar trend is observed when comparing F1 scores obtained from the logs mined with the Heuristics Miner (see Fig.~\ref{fig:process_level_metrics_road_traffic_finement}). In this setting, the finetuned Llama3.1-8B-Instruct and Ministral-8B-Instruct-2410 achieve F1 scores close to the original model (0.90), whereas the model derived from the finetuned Qwen2.5-7B-Instruct event log performs substantially worse, with an F1 score of 0.46. The differences in F1 scores between the Inductive and Heuristics Miners reflect the algorithms’ differing assumptions about process behavior, underscoring the importance of reporting results across multiple algorithmic approaches.

\begin{figure} 
    \centering 
    \includegraphics[width=0.85\linewidth]{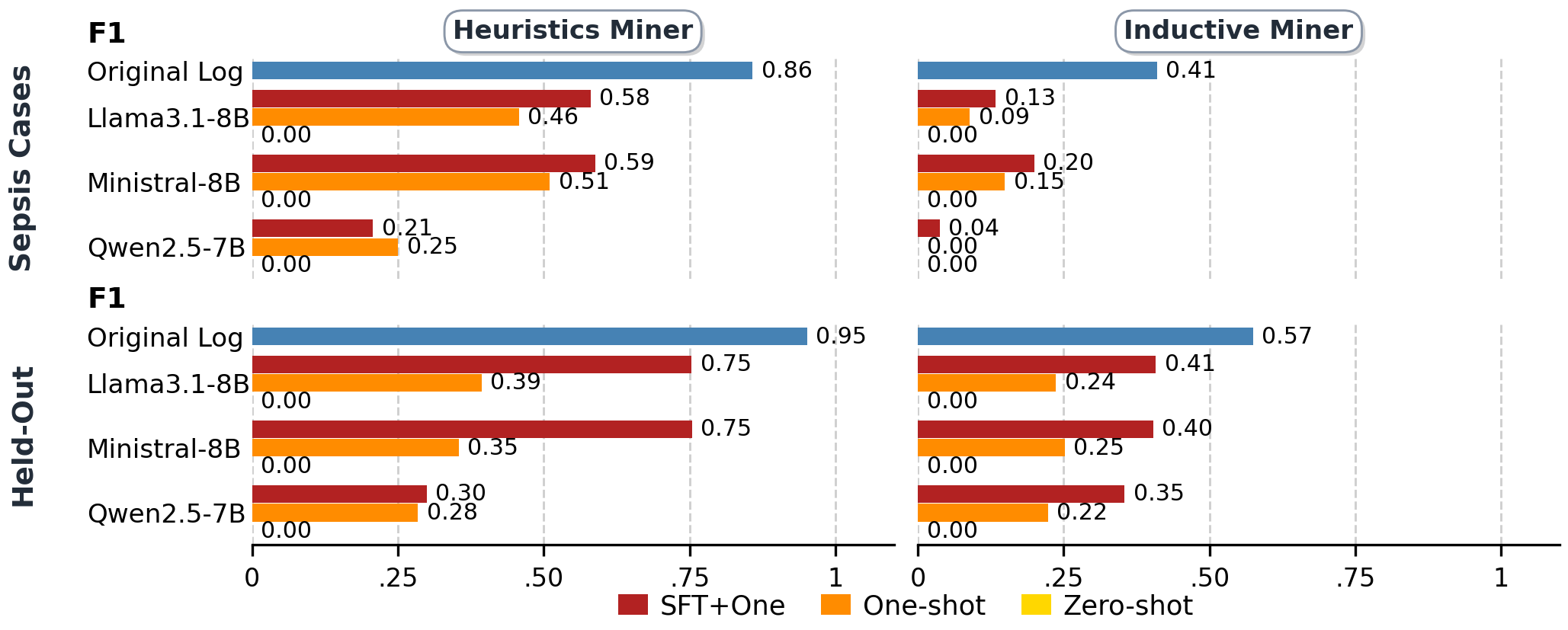} 
        \caption{F1 scores of Inductive and Heuristics process models from \textsc{Sepsis Cases} generated by the LLMs as well as from the held-out experiment with training on texts from GPT5.1 and evaluation on texts from Qwen3. On average, supervised finetuning (SFT) increases the event log quality from which the process models were derived, even if the training was performed on data from a different text distribution or from a different domain.}
\label{fig:process_level_ablation}
\end{figure}

To assess cross-domain and cross-distribution generalization, we mined process models from logs generated for synthetic discharge summaries and for texts from an unseen generator (Qwen3). Consider Fig.~\ref{fig:process_level_ablation} for an overview. Finetuning on police-report/XES pairs transfers to the medical domain: with the Inductive Miner, F1 improves by 4 pp. over the one-shot baseline of 0.09 for Llama3.1-8B-Instruct and from 0.15 to 0.20 for Ministral-8B-Instruct-2410, while Qwen2.5-7B-Instruct yields logs the miner can process at all only after finetuning. On \textsc{Sepsis Cases} with the Heuristics Miner, gains reach 12 pp. (Llama3.1-8B-Instruct) and 8 pp. (Ministral-8B-Instruct-2410); only for Qwen2.5-7B-Instruct is the one-shot setting stronger, as the Heuristics Miner better absorbs the high trace variance. Under the held-out-generator condition, F1 of the finetuned Llama3.1-8B-Instruct and Ministral-8B-Instruct-2410 almost doubles for heuristically mined models (+36 pp. and +40 pp.) and rises by 12–17 pp. for inductively mined ones for all three LLMs. The LLMs thus learn a generalizable text-to-log mapping, though the gap to the original-log values is larger than when all generators are seen during training, indicating some overfitting to the training text distribution.
\section{Discussion and Limitations}
\label{Discussion}

A primary consideration of this work is the use of synthetic corpora. Unlike real-world police documentation, which often features a 'form-and-narrative' hybrid structure \cite{martin2023learning}, our synthetic 'police reports' provide a different text distribution. We believe, however, that the semantic extraction capabilities required of the LLMs are largely invariant to these structural differences. Moreover, the utility of synthetic data generation should be viewed as a standalone contribution of the pipeline proposed in this paper: it offers a scalable method for finetuning models in specialized PM contexts where 'in-the-wild' data would otherwise be inaccessible.

Furthermore, our output is currently constrained to the XES standard. By adhering to this case-centric paradigm, the model does not yet account for the many-to-many relationships inherent in object-oriented structures, such as those defined in OCEL.
Extending this framework to support object-centric event data remains a priority for future iterations.

\section{Conclusion} 
\label{Conclusion}

We investigated the use of open-weight LLMs as direct translators from unstructured process descriptions to executable, standards-compliant event logs and process models. By generating paired synthetic text-log data from real-world process logs and using this data for supervised finetuning, we demonstrated that even relatively small LLMs can reliably produce faithful event data whose process representations closely resemble the original data. Our detailed evaluation at both the trace and process-model levels shows that finetuning substantially improves syntactic validity, reduces hallucinations, and yields data that supports meaningful process discovery with established mining algorithms. These results highlight supervised finetuning as a practical and scalable alternative to prompt or PM model only approaches using NLP techniques and LLMs, enabling better exploitation of previously untapped textual process descriptions.

Our findings suggest several promising directions for future research. An especially compelling avenue is the extension of this approach to training one LLM that can serve as a suitable 'data translator' across different domains, outputting event logs from different process descriptions. Using RL with verifiable rewards could also be helpful for out-of-distribution generalization capabilities. More broadly, integrating richer data modalities, exploring object-centric event representations, and validating the approach on real, non-synthetic reports remain important steps toward fully automated, end-to-end process data generation for PM practice.

\bibliographystyle{ACM-Reference-Format}
\bibliography{bibliography}




\end{document}